\documentclass[letterpaper]{article}

\newif\ifapx
\apxtrue %

\newcommand{\ourmaintitle}{Finding Interpretable Class-Specific Patterns through Efficient Neural Search}
\newcommand{\ourtitle}{\ourmaintitle}
\newcommand{\ourmethod}{\textsc{DiffNaps}\xspace}

\usepackage{subcaption}

\usepackage[arxiv]{aaai24}  %
\usepackage[notheorems]{eda}
\pgfplotsset{compat=1.9} 
\usepgfplotslibrary{fillbetween} 
\usepackage{times}  %
\usepackage{helvet}  %
\usepackage{courier}  %
\usepackage[hyphens]{url}  %
\usepackage{graphicx} %
\usepackage{natbib}  %
\usepackage{caption} %
\usepackage{algorithm}
\usepackage{algorithmic}
\usepackage{amsmath}
\usepackage{bbding}
\usepackage{pifont}
\usetikzlibrary{calc}

\usepackage{dsfont}
\usepackage{newfloat}
\usepackage{listings}
\DeclareCaptionStyle{ruled}{labelfont=normalfont,labelsep=colon,strut=off} %
\floatstyle{ruled}
\newfloat{listing}{tb}{lst}{}
\floatname{listing}{Listing}
\newcommand{\spumante}{\textsc{SPuManTe}\xspace}
\newcommand{\premise}{\textsc{Premise}\xspace}
\newcommand{\tree}{\cart}
\newcommand{\cart}{\textsc{Cart}\xspace}
\newcommand{\classy}{\textsc{Classy}\xspace}
\newcommand{\rrl}{\textsc{Rll}\xspace}

\newcommand{\Pro}{\mathds{P}}

\newcommand{\We}{W^E_b}

\newcommand{\barminmaxplot}[3]{
            \addplot+[forget plot, eda errorbarcolored, y dir=plus, y explicit]
    		table[x=ncols, y=#2_mean, y error expr=\thisrow{#2_max} - \thisrow{#2_mean}, col sep=comma] {./results/#1/#3_#1.csv};
    		\addplot+[eda errorbarcolored, y dir=minus, y explicit]
    		table[x=ncols, y=#2_mean, y error expr=\thisrow{#2_mean} - \thisrow{#2_min}, col sep=comma] {./results/#1/#3_#1.csv};

}

\newcommand{\barminmaxplotlimit}[4]{
            \addplot+[select coords between index={0}{#4},forget plot, eda errorbarcolored, y dir=plus, y explicit]
    		table[x=ncols, y=#2_mean, y error expr=\thisrow{#2_max} - \thisrow{#2_mean}, col sep=comma] {./results/#1/#3_#1.csv};
    		\addplot+[select coords between index={0}{#4},eda errorbarcolored, y dir=minus, y explicit]
    		table[x=ncols, y=#2_mean, y error expr=\thisrow{#2_mean} - \thisrow{#2_min}, col sep=comma] {./results/#1/#3_#1.csv};

}

\DeclareMathOperator*{\argmax}{arg\,max}

\usepackage{prettyplots}

\graphicspath{ {./figs/} }

\ifpdf
\usetikzlibrary{external}
\fi

\begin{document}
\usetikzlibrary{calc}
	\setlength{\pdfpagewidth}{8.5in}
	\setlength{\pdfpageheight}{11in}
	
	\title{\ourtitle}
	
	\author{
    Nils Philipp Walter\textsuperscript{\rm 1}, Jonas Fischer\textsuperscript{\rm 2}, Jilles Vreeken\textsuperscript{\rm 1}
	}
\affiliations{
    \textsuperscript{\rm 1} CISPA Helmholtz Center for Information Security,\\
    \textsuperscript{\rm 2} Harvard T.H. Chan School of Public Health, Department of Biostatistics\\
    nils.walter@cispa.de, jfischer@hsph.harvard.edu, vreeken@cispa.de

}
	
	\pdfinfo{
		/Title (\ourmaintitle)
		/Author (Nils Philipp Walter, )
		/Keywords (Interpretable Machine Learning, Information Theory)
	}	
	
	\maketitle
	
	\begin{abstract}
		
Discovering patterns in data that best describe the differences between classes allows to hypothesize and reason about class-specific mechanisms. 
In molecular biology, for example, this bears promise of advancing the understanding of cellular processes differing between tissues or diseases, which could lead to novel treatments. To be useful in practice, methods that tackle the problem of finding such \emph{differential} patterns have to be readily \textit{interpretable} by domain experts, and \textit{scalable} to the extremely high-dimensional data.

In this work, we propose a novel, inherently interpretable binary neural network architecture \ourmethod that extracts differential patterns from data. \ourmethod is scalable to hundreds of thousands of features and robust to noise, thus overcoming the limitations of current state-of-the-art methods in large-scale applications such as in biology.
We show on synthetic and real world data, including three biological applications, that, unlike its competitors, \ourmethod consistently yields accurate, succinct, and interpretable class descriptions. 

	\end{abstract}
	
	\section{Introduction}
\label{sec:intro}

Machine learning can be broadly categorized into \textit{predictive} and \textit{discovery}-based approaches.
\textit{Predictive} tasks, such as object detection, protein folding~\cite{alphafold} and fusion reactor control~\cite{tokamak}, are aimed at maximizing performance. Mastering such a given task often requires learning deep and intricate models from which it is hard up to impossible to understand how it arrived at a decision. 
In data-driven \textit{discovery}, the goal is to find \textit{interpretable} relations, called \textit{patterns}, in the data that best describe observed classes. That is, the focus is on interpretability rather than maximizing performance.
Discovery-based approaches are in especially high demand in biology, where the complex gene-regulatory dynamics and their differences between tissues or across diseases remain unclear, but, when elucidated, can offer new avenues for treatment and prevention.
Here, \textit{symbolic} explanations are essential for domain experts, for example, patterns of gene expression that are associated with cancer subtypes, to be able to directly understand and act on these patterns.

Although there exist massive amounts of high-dimensional data, such as genetic human variation or gene expression data, most existing approaches are not applicable as they either do not scale or are limited to pair-wise interactions. 
Here, we suggest a novel neural network learning approach that follows the paradigm of neuro-symbolic learning: leverage the predictive power of, and efficient frameworks for neural networks, while constraining the models such that learned patterns are fully interpretable.
In particular, we learn a modified NN architecture that in the forward pass leverages binary weights and activations to achieve symbolically interpretable intermediate features, while leveraging efficient continuous optimization during backpropagation~\cite{fischer2021differentiable}.

\begin{figure*}
    \centering
    \includegraphics[width=0.94\textwidth]{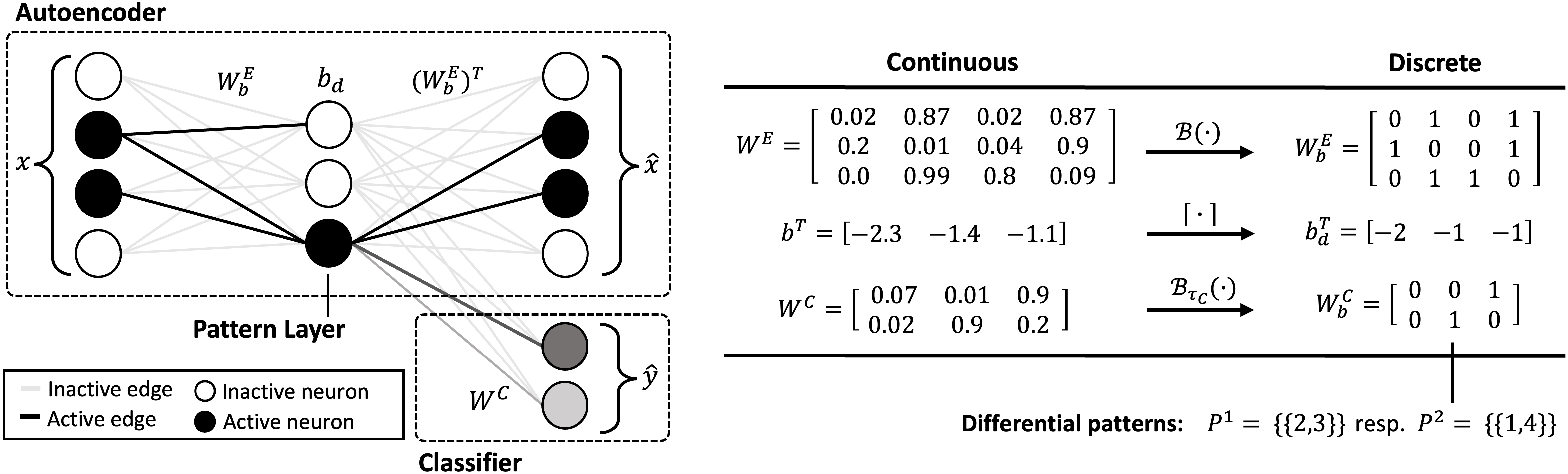}
    \caption{\textbf{Left:} The architecture of \ourmethod consists of a binarized autoencoder and a classifier  
    attached to the hidden layer. The neurons in the hidden layer encode patterns and are active if
    the corresponding pattern is present in the data.
    \textbf{Right:} The table shows the parameters of \ourmethod. In the forward pass,
    the continuous weights $W^E$ are stochastically binarized ($W^E_b$), while the classifier weights $W^C$ 
    are kept continuous. The bias $b$ in the hidden layer is ceiled to $b_d^T$. To
    extract the differential patterns (bottom right) per class  $P^1$ and  $P^2$, both matrices, $W^E$ and $W^C$,
    are deterministically binarized using the thresholds, $\tau_E$ and $\tau_C$. A pattern, encoded by a neuron,
    is given by the index set of all $1$ in the corresponding row of the weight matrix $W^E$. For the differential patterns,
    the binarized classifier weight matrix functions as a multiplexer to assign patterns to classes.%
    }
    \label{fig:diffnaps-arch}
\end{figure*}

To learn patterns that differentiate classes, such as healthy and tumor tissue, we build an architecture that is comprised of both a binary autoencoder and a separate classification head, which we call \ourmethod. We propose a multi-task objective to jointly optimize reconstruction and classification, driving learned patterns to differentiate between classes through a bottleneck in the autoencoder (see Fig.~\ref{fig:diffnaps-arch}). We additionally introduce regularizers that improve optimization and emphasize interpretability of learned patterns.

We empirically evaluate \ourmethod on synthetic and real-world data, comparing against baseline approaches such as classification trees, but also recent proposals such as rule lists, statistical and compression-based pattern mining, and neuro-symbolic learning. We show that \ourmethod faithfully reconstructs patterns relevant for distinguishing between classes, is robust to noise, and easily scales to hundreds of thousands of features, which makes it unique among existing work.
We consider three high-dimensional biological applications, including breast cancer genomics, on which \ourmethod finds meaningful patterns that hold promise for giving domain experts insight in the drivers of these diseases.

 	\section{Related Work}\label{sec:related}
Finding class-specific descriptions is at the core of discovery-oriented approaches in machine learning and data mining. A text-book example---and still widely used in practice---is the decision tree, which yields an interpretable decision path leading to a classification.

In data mining, \textit{emerging pattern mining} \cite{dong1999efficient,garcia:2018:overview} and \textit{subgroup discovery} \cite{kloesgen:95:sgd,atzmueller:2015:subgroup} are classic methods that aim to
discover the conditions under which the class labels assume an exceptional distribution.
Emerging pattern mining seeks to find \emph{every} such condition, which results in extremely many, highly redundant, and mostly spurious results.
Subgroup discovery yields the top-$k$ patterns with the strongest association with the target. While this circumvents the pattern explosion,
the results are still redundant \cite{vanLeeuwen:2012:diverse}. In contrast, we are interested in succinct and non-redundant descriptions.

\textit{Statistically significant pattern mining} \cite{llinares:2015:fast,pellegrina2019spumante} %
aims to discover patterns that have statistically significantly different distributions between classes.
These methods tend to suffer from the pattern explosion. That is, even on small data they often find tens of thousands redundant patterns, partially due to lack of multiple hypothesis test correction.

\textit{Pattern set mining} \cite{bringmann:07:chosen,budhathoki:15:diffnorm,hedderich2022label} solves this by asking for a non-redundant set of class-specific patterns that together describe the data well. These methods work well on small data, but as they are based on combinatorial-search heuristics that are (at least) quadratic in the number of features, they are mostly inapplicable to high-dimensional data.

\textit{Rule-based classification} \cite{lakkaraju:2016:interpretable,dash:2018:boolean,chen:2018:optimization,
proencca:2020:interpretable,hullermeier:2020:conformal,mctavish:2022:gosdt,huynh:2023:efficient,lin:2022:generalized}
aims to find interpretable classification rules of the form $\text{ if } X_1 = 1 \wedge X_5=1 \text{ then } Y=0$.
While such results are interpretable, these methods primarily focus on \textit{prediction} rather than \textit{description} and, hence, miss out on important details. 
Additionally, most are based on combinatorial optimization %
which prevents them from scaling to high-dimensional datasets.

\textit{Neuro-symbolic classification} \cite{wang:2020:transparent, wang:2021:scalable, kusters:2022:differentiable, dierckx:2023:rlnet}
has been proposed to overcome these computational limitations. These approaches design neural architectures from which, after training, symbolic classification rules can be extracted. Their optimization aside, in spirit these methods are similar to traditional rule-based classifiers as they focus on classification
accuracy rather than complete rule discovery. In contrast, \ourmethod combines data reconstruction with classification to discover the human-interpretable explanations relevant for the classes present in the dataset.

\section{Method}
\label{sec:method}
In this section, we introduce \ourmethod, a fully interpretable binary neural network-based approach
for finding patterns that describe the differences between classes in (very) high-dimensional data. 
We start by giving the intuition. 

\subsection{\ourmethod in a Nutshell}
Given a binary dataset and corresponding class labels, we seek to find interpretable patterns,
which succinctly and differentially describe the partitioning of the dataset induced by the labels.
That is, we want to find patterns that are more prevalent in a class than
in the rest of the data and, hence, allow  us to discriminate between classes.

To this end, we propose \ourmethod, a binary neural-network architecture designed
to find exactly such interpretable patterns. The architecture consists of a two-layer binary autoencoder, combined with a classification head (see Figure \ref{fig:diffnaps-arch}).
The classification head is a fully connected layer, attached to the hidden layer of the autoencoder.

During the forward pass, we interpret the continuous weights $w_{ij}\in[0,1]$ in the autoencoder as Bernoulli variables distributed as
$\mathcal{B}(w_{ij})$ and binarize them stochastically. Each neuron performs a dot product between the binary weights and input. The neuron is active if the 1s in the input align with the 1s in the weight vector. Thus the weight vector can be interpreted as a pattern and hence we refer to the hidden layer as the \emph{pattern layer}. Intuitively, the weights are optimized such that the autoencoder---the set of encoded patterns---reconstructs the data well.

To find differential patterns, we need to reward those patterns that are specific for a class. 
We achieve this by adding a classification head, corresponding to a logistic regression on the pattern layer.
That is, we seek to classify samples based on the presence and absence of patterns.

To find a good set of patterns, the network is trained using a multi-task loss.
The autoencoder is trained to  minimize the reconstruction error, while the classification head is trained
to minimize the classification error. As such, the network is driven towards
the learning of relevant patterns in the data that are at the same time differential between classes.

\subsection{\ourmethod in Detail}

Next, we discuss \ourmethod in detail. We first introduce notation, and then, in turn, discuss the architecture, how to extract differential patterns, how to carry out the forward pass, the multitask loss, and how to backpropagate errors through \ourmethod.

\subsubsection{Notation}
We consider labeled binary datasets $(X,Y) \in\{0,1\}^{n \times m} \times \{1,...,K\}^n$ of $n$ samples, $m$ features and $K$ classes.
We write $X_{i,j}$ to refer to the value of the $j$-th feature of the $i$-th sample.  
We denote the partition of the dataset for class $k$ by $X^k = \{X_i\mid Y_i=k\}$.

A pattern $p$ is a subset of feature indices $p\subseteq \{1...m\}$ and represents feature co-occurrences.
A row $X_i$ contains a pattern $p$ iff $X_{ij}=1, \, \forall j \in p$. 
The support $\mathit{supp}(p)$ of a pattern $p$ is the number of rows that contain $p$, 
and analogue $\mathit{supp}_k(p)$ is the support where additionally $Y_i = k$.
We have 
\begin{align}
    \Pro(p \mid k) = \frac{\mathit{supp}_k(p) }{n_k} 
    \quad \text{and} \quad
    \Pro(k \mid p ) = \frac{\mathit{supp}_k(p) }{\mathit{supp}(p)} 
    \; ,
\end{align}
where $n_k$ is the number of samples where $Y_i = k$. %

We say a pattern $p_k$ is \emph{differential} for class $k$ if it both has a higher support in $X^k$ than
in $X \setminus X^k$, and the probability of class $k$ is highest for records that contain $p_k$. Formally, iff
$$ k =\argmax_{k' \in \{1...K\}} \Pro(p_k \mid k' ) = \argmax_{k' \in \{1...K\}} \Pro(k' \mid  p_k   )\; .$$
Our goal is to find a set $P^k$ of such patterns per class $k$. 

\subsubsection{Architecture}
The architecture of \ourmethod consists of a binary autoencoder and a classification head attached to the hidden layer. We graphically depict it in Fig. \ref{fig:diffnaps-arch}.

The encoding and decoding layers of the autoencoder share a set of continuous weights $W^E$, which are learned during backpropagation.
The forward pass uses a binarized version of this weight matrix $W^E_b$.
A hidden neuron $j$ represents a pattern, and a feature $i$ is part of the pattern corresponding to neuron $j$, iff $W^E_b[i,j] = 1$.
The activation function of the encoder $\lambda_{E}$ is a binary step function centered at a learned bias term, which represents how many features need to be present for the neuron to "fire"---i.e. for the pattern to be considered present in the sample. We refer to the hidden layer as the pattern layer.

The decoding layer performs the transposed linear transformation of the encoding layer i.e. $W^d_b = (W^E_b)^T$. 
Hence, if a neuron is active, the pattern encoded in that  neuron is used as a whole for the reconstruction.
Consequentially, to achieve a low reconstruction loss,  the patterns formed during optimization must succinctly describe the data.

To reward differential patterns, we connect a classifier to the pattern layer with continuous weights $W^C$ that is tasked to predict the label of a sample based on the presence and absence of patterns. The classifier is linear, and, hence, highly interpretable.
To extract differential patterns, we binarize weight matrices $W^E$ and $W^C$ by thresholding with $\tau_e$ and $\tau_c$, respectively. As described above, the patterns in the pattern layer are given by the index set of all $i$'s such that $W^E_b[i,j] = 1$. 
The discretized classifier weights allow us to assign patterns to their respective classes.
For a formal description of the pattern extraction, we refer to App.~\ref{app:extracting}.

\subsubsection{Forward Pass}
\label{sec:forward}
We denote the size of the hidden dimension of the autoencoder by $h$ and the binary weights of the encoder as
$W^E_b \in \{0,1\}^{h\times m} $.  We define a linear layer without bias as $ f_{W}(x) = Wx.$
For a binary input $x\in\{0,1\}^m$, we compute the activations of the pattern layer as
$$z = f_{E} (x)=\lambda_{E}(f_{W^E_b}(x)) \; .$$
where $\lambda_E: \mathds{R} \rightarrow \{0,1\}$ is the binary step function as defined by \citet{fischer2021differentiable}.
To steer the encoded patterns to be differential rather than merely descriptive, we attach a classifier to the pattern layer.
This classifier has continuous weights $W^C \in [0,1]^{K\times h}$ and computes a linear transformation followed by a softmax of the binary hidden activations $\hat{y} = \operatorname{softmax}(f_{W^C}(z))$. That is, its output depends only on the presence or absence of patterns.

To ensure interpretability, we use the transposed encoder weights as weights of the decoder $W^D_b=(W^E_b)^T \in \{0,1\}^{m\times h} $.
The  reconstruction $\hat{x}$ of the input $x$ is given by
$$ \hat{x} = f_{D} (x)=\lambda_{D}(f_{W^D_b}(z)) \; ,$$
where $\lambda_{D}$ is the activation of the decoder as defined by \citet{fischer2021differentiable}, clamping the input to the interval $[0,1]$ and rounding it to the closest integer.

\subsubsection{Objective Function}
Our objective function consists of four terms: one for the autoencoder, one for the classification, and two regularization terms.
To optimize the classifier, we use the cross-entropy loss between the predicted logits $\hat{y} $ and the one-hot encoding
of the ground truth label $y$:
$ l_c(y,\hat{y}) = \sum_{k=1}^{K} y_k \operatorname{log}(\hat{y}_k).$
As binary tabular data tends to be sparse, i.e., the number of ones \#1 and number of zeros \#0 are highly unbalanced, 
we use a sparsity-aware reconstruction loss~\citep{fischer2021differentiable}
that weighs the importance of reconstructing a 1 proportional to the sparsity of the data. 
For a sample $x \in \{ 0,1\}^m$ and reconstruction $\hat{x} \in \{ 0,1\}^m$, the reconstruction loss is 
$$ l_e(x,\hat{x}) = \sum_{j=1}^m ((1-x_j)\alpha + x_j (1-\alpha)) | x_j - \hat{x}_j | \; , $$
where $\alpha = \frac{\#1s}{\#1s + \#0s}$ is the sparsity of the data. 

Our overall goal is to find a succinct description of the classes in terms of  
class-specific patterns encoded by the neurons in the hidden layer.
To promote such patterns, we adapt the $\mathcal{L}_2$-regularizer to penalize long patterns i.e. rows with a lot of $1$s. This adapted regularizer is given by
$$r_s(W) = \sum_{i=1}^m \left(\sum_{j=1}^{h} W_{i,j}\right)^2 \; .$$
Instead of considering each weight individually, we sum the rows before squaring them. This penalizes a pattern as a whole by imposing
a quadratic cost on the length of the pattern.
Hence, the regularizer tilts the optimization to prefer shorter patterns. To further push the weights to a binary solution we  employ a W-shaped regularizer
\cite{bai_proxquant_2019,dalleiger2022efficiently}, defined as
$$r_{b}(W)=\sum_{w \in W} \min \{r(w), r(w-1)\}\; ,$$
$$r(w)=\kappa\|w\|_1+\lambda\|w\|_2^2 \; .$$
This regularizer is based on the elastic-net regularizer and the hyperparameters $\kappa$ and $\lambda$ specify the
trade-off between the ridge and lasso penalty. For $\kappa = \lambda = 1$, the regularizer is depicted in Figure \ref{fig:regu} in the Appendix.
Compared to $r_s$, the W-shape regularizer is applied element-wise to push the individual weights towards zero or one.

In the forward pass, we apply stochastic quantization $W_b[i,j]\sim \mathcal{B}(W[i,j] )$. If all $W[i,j],\text{for }j=\{1...m\}$,
have the same value, a sample of a row is binomially distributed with $p=W[i,j]$ and $m$ trials. 
The expected value is then $mW[i,j]$. Considering a minimum of two features for a neuron to fire, this means that when all $W[i,j]$ drop below $1/m$ the neuron is on expectation `dead'. To prevent regularizers from zeroing out a neuron by pushing $W_{ij}$ below this threshold, we offset the weights by $-1/m$ before applying the regularizers.
For the same reason, we set the gradients for $r_s$ to zero if $\,\sum_{j=1}^{h} W[i,j] < 1\;.$

Given the parameters of the network $\theta = \{W^E,W^C\}$ the loss function 
for a dataset $(X,Y)$ %
is given by
$$\mathcal{L}(X,Y;\theta) = \sum_{i=1}^{n} l_e(y_i,\hat{y}_i) + \lambda_c \, l_c(x_i,\hat{x}_i) + r_s(W^E)+r_b(\theta)\;,$$
where $\lambda_c$ is a parameter that weighs the classification loss.

\subsubsection{Backward Pass}
\label{sec:backpass}
We  minimize this loss function using gradient descent. 
For this, we need to compute the partial derivatives with respect to the weights of the network.
To be able to pass gradients through step-functions, we use the straight-through-estimator (STE), which is commonly employed in binary neural networks~\cite{ste}. For a particular layer, $g_u$ denotes
the upstream gradient. For the derivatives with respect to the autoencoder, we follow the approach of 
\citet{fischer2021differentiable}. In particular, for encoding layer $W^E$ and input $x$

$$\frac{\partial f_{\We}}{\partial W^E}:=g_u x^{\top} \; , \quad \frac{\partial f_{\We}}{d x}:= (W^E)^{\top} g_u \; .$$

The derivative through the activation function of the decoder $\lambda_D$ is given by
$\frac{\partial \lambda_D}{\partial x}:=\mathds{1}g_u$. For the activation 
function of the pattern layer, the STE above is inapplicable.
In the case that features are wrongly reconstructed, the resulting loss  would propagate negative
gradients through the STE, even to inactive neurons. Hence, we adapt the \emph{gated} STE,
which gates the gradient depending on whether a neuron was active in the forward pass. The derivatives
for bias $b$ and input $x$ are 

$$
\frac{\partial \lambda_E}{\partial b}:=\begin{cases}
g_u & \text { if } \lambda_E(x)=1 \\
0 & \text { if } \lambda_E(x)=0
\end{cases} ,
$$

$$
\frac{\partial \lambda_E}{\partial x}:= \begin{cases}g_u & \text { if } \lambda_E(x)=1 \\
\max \left(0, g_u\right) & \text { if } \lambda_E(x)=0\end{cases}.
$$
In quantized neural networks, it has been observed that quantizing the classification layer has a negative impact on performance
\cite{choi2018bridging, liu2018bi,hubara2017quantized}. Thus we do not quantize the weights of the classifier  during training.
Although the classifier is not quantized, the classifications
are transparent and interpretable, since the classification head is similar to logistic regression and the weights
are constrained to be in the interval $[0,1]$.

Finally, after a round of backpropagation, all weights
are clipped to the interval $[0,1]$. This enables stochastic binarization for the autoencoder and
the classifier for the next forward pass and to transparently interpret the contribution of a pattern to a certain class.
We clamp the bias at a maximum of $-1$, such that at least two features have to be present
for a neuron to become active. 

This concludes the formal description of \ourmethod.

\subsubsection{\ourmethod in Practice}
To use \ourmethod in practice, we need to choose the number $h$ of hidden neurons and set $\lambda_c$. 

For medium to high-dimensional data, setting the size of the hidden layer lower than the dimensionality of the data, $m$, creates an inductive bias towards differential patterns. Since to achieve both a low reconstruction loss and low classification loss, the patterns in the hidden layer have to be predictive, i.e., high $\Pro(k\mid p_k)$, and due to the bottleneck, the patterns must cover the partition well, i.e., high $\Pro(p_k\mid k)$. %

For low dimensional data, choosing a small hidden layer results in an under-parameterized 
network that will underfit. Choosing a larger hidden layer, thus having more parameters, 
outweighs the benefits of the bottleneck.

Parameter $\lambda_c$ weighs the effect of the reconstruction and classification losses. 
The magnitude of the reconstruction loss varies strongly among different datasets.
In practice, we increase $\lambda_c$ until the classification error saturates.

	\section{Experiments}
\label{sec:exps}

We compare \ourmethod five state-of-the art methods on synthetic and real-world data. In particular, we compare to
decision trees \cite[\cart,][]{breiman2017classification},
significant pattern mining \cite[\spumante][]{pellegrina2019spumante}, 
MDL-based label-descriptive \cite[\premise,][]{hedderich2022label} and
classification rule learning \cite[\classy,][]{proencca:2020:interpretable}, and
neuro-symbolic classification rule learning \cite[\rrl,][]{wang:2021:scalable}.

We additionally considered top-$k$ subgroup discovery \cite{lemmerich2018pysubgroup}, difference description \cite{budhathoki:15:diffnorm},
falling rule lists \cite{chen:2018:optimization,lin:2022:generalized}, optimal sparse decision trees~\cite{mctavish:2022:gosdt}, and class-specific BMF \cite{hess2017c}, but found these do not scale to, or do not find patterns on non-trivial data.

\premise and \spumante consider only binary classes. To allow fair comparison in a multiclass scenario,
we run them in a one-versus-all for each class and merge the results.

The hyperparameters for the predictive approaches are tuned based on accuracy on a hold-out set.
For \spumante, we used the default parameters given by the authors. We fit the
hyperparameters of \ourmethod based on our loss function.
The experiments for the neural approaches, i.e. \ourmethod and \rrl, are executed on GPUs. 
For more on the experimental setup, we refer to Appendix~\ref{app:expdetails}.

\subsection{Synthetic Data}
To evaluate all methods on data with known ground we first consider synthetic data. 
We measure success in terms of soft F1~\cite{hedderich2022label}, by which we avoid overly penalizing methods that recover only parts rather than exact matches of ground truth patterns. 
The formal definition can be found in Appendix \ref{app:f1}.
Informally, the soft F1 score does not require strict equality between a discovered pattern
$p_d$ and the corresponding ground truth pattern $p_g$ but uses a soft equality, i.e., the Jaccard
distance of $p_d$  and $p_g$.

\subsubsection{Data Generation}
In the experiments below we generate synthetic data as follows. We start with an empty data matrix of $n$ rows and $m$ features.
We sample $10$ patterns per class, uniformly at random (u.a.r.) across features, drawing their length from $\mathcal{U}(5,15)$. 
We sample $20$ common patterns u.a.r, but draw their length from $\mathcal{U}(0.01m,0.025m)$ to maintain the density of the data. 
Per class, we generate equally many rows.
Per row, we plant u.a.r. two common and three class-specific patterns. We then apply both additive noise by flipping ten $0$s to $1$s, as well as destructive noise by flipping $1$s due to a pattern to $0$s with a probability of $2.5\%$. 
Finally, we assign the class label such that $\mathds{P}( k \mid p \in P_k)=0.9$. 
Unless specified otherwise, we report the average results over five independently drawn datasets.

\subsubsection*{Scalability in $m$} \label{sec:scam} First, we consider how well \ourmethod scales to high dimensional data. 
We fix the number of classes K to $2$, the number of rows $n$ to $10\,000$, and vary $m \in \{100, 500, 1\mathrm{k}, 5\mathrm{k}, 10\mathrm{k}, 15\mathrm{k}, 20\mathrm{k}, 25\mathrm{k}, 50\mathrm{k}, 100\mathrm{k}\}$. 
To reduce the overlap across patterns in low-dim. data $m < 1000$, we sample 5 patterns per class and no shared patterns. 

\begin{figure*}
    \centering
    \begin{subfigure}[b]{0.25\textwidth}
            \begin{tikzpicture}
        \usetikzlibrary{calc}
        \begin{loglogaxis}[ 
            small line,
            legend style={xshift=4.25cm,yshift=0.65cm},
            width = \linewidth,
            height=3cm,
            ylabel={time (min)},
            xlabel={Number of features $|m|$},
            ymax=800.0,
            legend columns = 7,
            tick label style={/pgf/number format/fixed},
            y tick label style 	= {font=\scriptsize, xshift = 1pt},
            y label style 		= {at={(axis description cs:-0.275,0.5)}, anchor=south, font=\scriptsize},
        ]
        \foreach \x in {diffnaps, tree,  mdlrl, spumante, premise, rrl}{
               \barminmaxplot{\x }{time}{exp1};
        }
        \end{loglogaxis}
      \end{tikzpicture}
      \subcaption{Runtime (lower is better)}
      \label{fig:exp1a}
    \end{subfigure}%
    \begin{subfigure}[b]{0.25\textwidth}
    \begin{tikzpicture}
        \usetikzlibrary{calc}
        \begin{semilogxaxis}[ 
            small line,
            legend style={xshift=4.35cm,yshift=0.65cm},
            width = \linewidth,
            height=3cm,
            ylabel={SoftF1},
            xlabel={Number of features $|m|$},
            ymax=1.0,
            legend columns = 7,
            xmode = log,
            xmax=100000,
            tick label style={/pgf/number format/fixed}
        ] %
        \foreach \x in {diffnaps, tree,  mdlrl, spumante, premise, rrl}{
               \barminmaxplot{\x }{F1}{exp1};
        }      
            \addlegendentry{\ourmethod}
            \addlegendentry{\cart}
            \addlegendentry{\classy}
            \addlegendentry{\spumante} 
            \addlegendentry{\premise}
            \addlegendentry{\rrl}
        \end{semilogxaxis}
      \end{tikzpicture}
      \subcaption{F1-score (higher is better)}
      \label{fig:exp1b}
    \end{subfigure}%
    \begin{subfigure}[b]{0.25\textwidth}
    \begin{tikzpicture}
        \usetikzlibrary{calc}
        \begin{axis}[ 
            small line,
            legend style={xshift=4.25cm,yshift=0.65cm},
            width = \linewidth,
            height=3cm,
            ylabel={SoftF1},
            xlabel={Number of classes $K$},
            ymax=1.0,
            legend columns = 6,
            tick label style={/pgf/number format/fixed}
        ]

        \foreach \x in {diffnaps, tree,  mdlrl, spumante, premise, rrl}{
               \barminmaxplotlimit{\x }{F1}{exp2}{10};
        }
        \end{axis}
      \end{tikzpicture}
      \subcaption{F1-score (higher is better)}
      \label{fig:exp2}
    \end{subfigure}%
    \begin{subfigure}[b]{0.25\textwidth}
    \begin{tikzpicture}
        \usetikzlibrary{calc}
        \begin{axis}[ 
            small line,
            legend style={at={(0.45, 1.1)}, anchor=south},
            width = \linewidth,
            height=3cm,
            ylabel={SoftF1},
            xlabel={Destructive noise},
            ymax=1.0,
            legend columns = 6,
            tick label style={/pgf/number format/fixed}
        ]

        \foreach \x in {diffnaps, tree,  mdlrl, spumante, premise, rrl}{
               \barminmaxplot{\x }{F1}{exp4};
        }
 
        \end{axis}
      \end{tikzpicture}
      \subcaption{F1-score (higher is better)}
      \label{fig:exp4}
    \end{subfigure}
      \caption{ \textit{Scalability.}
      We show runtime \textbf{(a)} and F1-scores \textbf{(b)} when varying $m$, resp. F1-scores when varying the number of classes $K$ \textbf{(c)} and varying the amount of destructive noise \textbf{(d)}. 
      Our method, \ourmethod can confidently handle a large number of features with a negligible increase in runtime, outperforms all state-of-the-art competitors significantly in terms of F1.
      }\label{fig:synth}
\end{figure*}
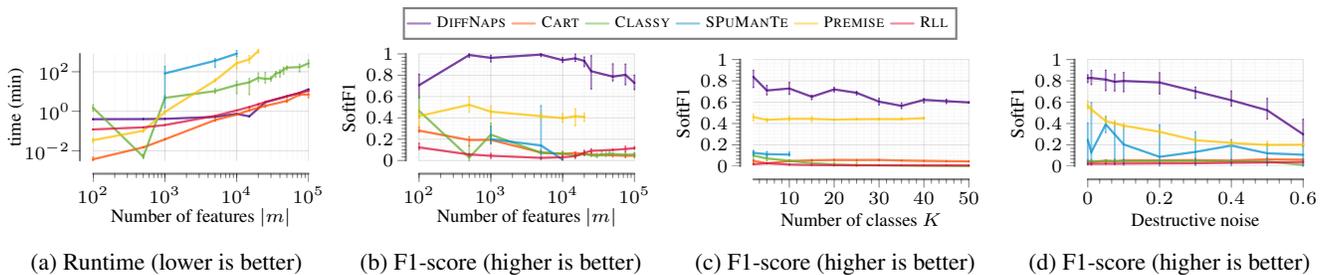

We run all methods and report their  results in Fig.~\ref{fig:synth}a,b. 
Except for \premise and \spumante, all terminate within 24 hours. 
\premise runs out of time for $m>20\mathrm{k}$.  \spumante runs out of memory for $m<1\textrm{k}$ and $m>10\textrm{k}$. 
We see in Fig.~\ref{fig:exp1a} that \classy is one order of magnitude slower than \ourmethod, \rrl, and \tree all of which perform on par in terms of runtime. %

Next, we inspect the average F1-scores, which we show in Fig.~\ref{fig:exp1a}. 
We see that \classy, \rrl and \tree perform poorly, as they recover only small parts of the ground truth patterns, 
and that \spumante varies in performance due to having to sub-sample the data.
\premise achieves scores of  approx. $0.5$ across all $m$.
\ourmethod consistently recovers ground truth well across many orders of magnitudes of $m$. 
For very high dimensional data,  performance slowly deteriorates but still outcompetes the state-of-the-art by a wide margin.

\newcommand{\auc}{{\textsc{auc}}}
\newcommand{\na}{{--}}
\newcommand{\colspacingstats}{\hspace{1em}}
\newcommand{\colspacingmethods}{\hspace{2em}}
\newcommand{\colspacingresults}{\hspace{1em}}

\begin{table*}[!ht]
    \centering \fontsize{9}{11}\selectfont 
    \begin{tabular}{ l rrr rrr rrr rrr rrr rrr} 
        \toprule
         & &  &  & \multicolumn{3}{c}{\ourmethod (ours)}   & \multicolumn{3}{c}{\cart} & \multicolumn{3}{c}{\classy} &\multicolumn{3}{c}{\premise} & \multicolumn{3}{c}{\spumante}\\
         \cmidrule(lr){5-7} \cmidrule(r){8-10} \cmidrule(r){11-13} \cmidrule(r){14-16} \cmidrule(r){17-19}
        Dataset & $n$ & $m$ & $K$  & $\#P$ & $\overline{|P|}$ & \auc & $\#P$ & $\overline{|P|}$ & \auc & $\#P$ & $\overline{|P|}$ & \auc & $\#P$ & $\overline{|P|}$ & \auc & $\#P$ & $\overline{|P|}$ & \auc      \\
        \midrule
        Cardio     & 68k & 45     & 2  & 14        & 2 & .56 & 7k            & 6 & \textbf{.62} & 10 & 2 & .36 & 28 & 1 & .51 & 346         & 4 & .56 \\
        Disease  & 5k  & 131    & 41 & 838         & 2 & \textbf{.84} & 1    & 2 & .00 & 25 & 2 & .11 & 187         & 3 & \textbf{.84} & 2k & 3 & .39\\
        BRCA-N            & 222   & 20k  & 2 & 146 & 9 & .91 & 1            & 2 & .00 & 3 & 1 & .45 & \na & \na & \na & 4k                  & 3 & \textbf{.95}\\
        BRCA-S            & 187   & 20k  & 4 & 1k  & 2 & \textbf{.86} & 22  & 2 & .31 & 2 & 1 & .23 & \na & \na & \na & 0 & 0 & 0\\
        Genomes       & 2.5k  & 225k & 6 & 732     & 7 & \textbf{.77} & 127 & 4 & .46 & 7 & 2 & .36 & \na & \na & \na & \na & \na & \na\\
        \bottomrule
    \end{tabular}
    \caption{\textit{Real world data} We give the number of samples ($n$), features ($m$), and classes ($K$), and per method the number of discovered patterns ($\#P$), average length of discovered patterns ($\overline{|P|})$ and area under the curve (AUC) of the sensitivity-specificity plot (see App.~Fig.~\ref{fig:odd-coverage}, Sec~\ref{app:auc}). We aborted experiments taking longer than 24h or running out of memory (\na). %
    }
    \label{tab:real}
\end{table*}
    
\subsubsection{Multi-class} Next, we examine how well \ourmethod scales to a large number of classes. 
To this end, we generate data as above, varying the number of classes $K \in$  $\{2,5,10,15,...,25,50\}$, generating $1\,000$ rows per class, setting $m = 5\,000$. We give the results in Fig.~\ref{fig:exp2}. 
\rrl, \classy, \cart, and \spumante fail to recover more than a small subset of the ground truth. In addition \spumante runs out of memory for more than $10$ classes.
\premise achieves scores of around $50\%$ up to 20 classes, but fails to terminate for $K\geq 40$, as running in a one-versus-all setting incurs high computational costs. In contrast, \ourmethod stably performs best in this  setting.

\subsubsection{Robustness to Noise} 
Finally, we evaluate how robust methods are to noise. 
Here, we set $K=2$, $m=5\,000$, and $n=2\,000$.
First, we consider additive noise by varying the number of randomly added $1$s per row, from $0$ to $100$. 
In the interest of space, we postpone the Figure to App.~Fig.~\ref{app:fig-noise}.
We find that \spumante  rapidly fails to discover meaningful results, and runs out of memory for $a$ = $100$. 
In contrast, \ourmethod and \premise are robust across varying $a$, with \ourmethod outperforming its competitors by a wide margin.

Second, we consider destructive noise by varying the probability of flipping a $1$ to a $0$, from $0\%$ to $60\%$.
We show the results in Fig.~\ref{fig:exp4}. 
\tree, \classy, and \rrl all obtain F1 scores of near-zero, \spumante performs slightly better on average but shows a large variance in the performance across repetitions.
\premise is the best among competitors, but its performance declines rapidly even for small amounts of destructive noise. 
In contrast, \ourmethod is robust, its performance virtually unaffected up to $20\%$ destructive noise, i.e., up to a signal-to-noise ratio of 6dB.  

\subsection{Real-World Data}

Next, we evaluate \ourmethod on five biological datasets. We consider phenotypical \emph{Cardio} data~\cite{cardio},
a \textit{Disease} diagnosis~\cite{disease} dataset, 
two high-dimensional binarized gene expression datasets for breast cancer, \textit{BRCA-N} and \textit{BRCA-S}, that we derived from The Cancer Genome Atlas (TCGA) (see App.~\ref{app:brca-gen}), 
and a human genetic variation data set~\cite{many:15:1kgenomes, fischer:20:mexican}.

We consider the same competitors as before, except \rrl as it returns no patterns for any data but \emph{Cardio}. 
To obtain results with \spumante we had to restrict it to 250 samples for \emph{Cardio}, 4000 for \emph{Disease}, 50 for both \emph{BRCA} datasets. We could not find any setting to make it work on \emph{Genomes}. We report running time for all methods in App.~Tab.~\ref{app:tab-runtime}.

\subsubsection{Quantitative Results}
As the ground truth is unknown, we report the number of discovered patterns, their average length, and the area under the curve of 
what percentage of the data the patterns cover when we order them by probability of seeing a class given a pattern 
(see App.~\ref{app:auc}). Intuitively, this corresponds to sensitivity (how much do we cover) versus specificity (how specific are patterns for that class). To filter spurious patterns we compute this measure over patterns for which at least $1/k$ + $0.1$ of their probability mass is assigned to one class. For example, for the binary setting only those patterns that occur at least $60\%$ in the class, where $50\%$ would correspond to independence (random coin flip). 

We report basic statistics and results in Tab.~\ref{tab:real}. 
We observe that \ourmethod performs well across all datasets, obtaining AUC scores that are either best by a wide margin (\textit{Disease}, \emph{BRCA-S}, \textit{Genomes}) or close second best (\textit{Cardio}, \textit{BRCA-N}). %
Consistent with our synthetic data study, our competitors yield mixed results; they do not scale to high dimensional data (\premise, \spumante), or return prohibitively many or unspecific patterns (\spumante, \tree, resp. \classy). 

Regarding the length of discovered patterns, we observe that those by \ourmethod reflect the 
complexity of the datasets: on \textit{Cardio} and \textit{Disease}, which contain complex, information rich features, it finds smaller patterns, while for the other datasets, that consist of low-level molecular information as features, it finds longer patterns to capture complex relationships. 
In contrast, \classy generally discovers only few, medium-length patterns across datasets, while \tree recovers more complex relationships that are, however, less descriptive of the classes as measured by the AUC.

\subsubsection{Qualitative Results}
Next, we analyse the results of \ourmethod in detail and show their relevance for biological research on the breast cancer datasets.\!\footnote{Human variation data, such as the \textit{Genomes} dataset, is an ideal application for \ourmethod as it is a high-dimensional resource of binary data in which we can uncover potential genetic predispositions of individuals to diseases, thus allowing to advance early detection and treatment. However, in the available data, the target class is the population membership of the individual, which raises ethical concerns for detailed analysis. Sadly, no further meta-data is available to meaningfully split \textit{Genomes} for differential analysis.} 

\vspace{0.5em}
\noindent \emph{Differentiating Breast Cancer and Healthy Tissue} Breast cancer (BRCA) is the most common cancer and the leading cause of death from cancer among women in the world~\cite{brca-review}. The exact underlying gene regulatory dynamics are actively researched.

We apply \ourmethod on \emph{BRCA-N} and discover $146$ differential patterns of gene co-expression for BRCA and adjacent normal tissue. To see if these capture relevant molecular differences, we run a statistical gene set over-representation analysis using KEGG (see App.~\ref{app:brca-ana}), a manually curated gold standard for molecular interactions, reactions, and relations~\cite{kegg}. 

We first do a pooled analysis over all genes identified by any discovered pattern for a class, i.e., the union of features in the respective patterns. 
We find that enriched pathways for tumor tissue correspond to known cancer drivers, such as MAPK and WNT signalling, while for the healthy tissue we find pathways linked to the regulation of lipolysis in adipocytes as well as PPAR signalling, both of which are known to be dysregulated in BRCA~\cite{brca-adipocyte, brca-ppar}. In short, \ourmethod discovers patterns that together describe complex, cancer-related functions. 

Investigating \textit{individual} patterns, we find that while many identify general pathways like above, others are enriched for \textit{specific} pathways, such as PPAR. This shows the discovered patterns reveal details that can potentially be used for discovering alternative treatment targets for these pathways.

\vspace{0.5em}
\noindent \emph{Differentiating Cancer Subtypes} It is well known that breast cancer is not one single disease, e.g. the Luminal A, Luminal B, HER2+, and the Triple Negative subtypes all show distinct molecular behaviour, response to treatment, and patient survival.
To investigate whether \ourmethod can elucidate differences between these subtypes, we run \ourmethod on \emph{BRCA-S}, a balanced dataset of primary BRCA tissue with subtype label, and again analyse the discovered patterns using a gene set over-representation analysis in KEGG.

Starting with a pooled analysis, we find significantly enriched pathways that capture specifics of classes.
Luminal A, for example, is defined by a lack of HER2. For this subtype, \ourmethod discovers patterns that are enriched for (i.e. related to) dilated cardiomyopathy. This is a common side-effect in Trastuzumab treatment, a drug which targets and depletes HER2 in HER2 positive subtypes~\cite{brca-trastu}. 
Luminal B is Estrogen receptor positive, meaning it expresses this receptor.
For this subtype, we find patterns that are significantly enriched for sphingolipid metabolism.
This is an important component for cell survival, proliferation, and promotion of cell migration and invasion in Estrogen receptor-positive BRCAs~\cite{brca-sphingo}.
These metabolites are also targets of treatment, and the discovered patterns could reveal insights leading to potential new therapeutic targets.

\vspace{0.5em}
\noindent \emph{Promising Novel Patterns} 
On both \emph{BRCA} datasets, we find highly class-specific patterns, with average log-odds of $\Pro(p\mid k)$ against $\Pro(p\mid \lnot k)$  of $\approx8$ resp.  
$\approx 5$. Encouragingly, the above analysis above showed that many of these patterns capture complex biological processes related with BRCA progression or tumorigenes. More exciting perhaps are those patterns for which the genes are not yet annotated in a pathway but are strongly associated with BRCA or its subtypes. We are looking forward to conducting an in-depth analysis with oncologists, relating these patterns with more fine-grained subtypes or treatment groups.

	\section{Discussion}
\label{sec:discussion}

Experiments show that \ourmethod finds succinct sets of differential patterns, 
scales to hundreds of thousands of features, large number of classes, and is robust to noise.

On synthetic data, we saw  that existing methods fail to recover significant portions of the planted differential patterns. 
Rule-based methods only recovered small subsets of incomplete patterns. 
\spumante suffers from memory problems, and
returns  overly large, redundant results. 
\premise does account for redundancy, which results in better performance, but its combinatorial search does not scale well.
\rrl and \tree scale very well, but show poor performance on synthetic data.
Surprisingly, none of the existing approaches are robust to destructive noise. 

On real world data, we find \ourmethod is the only approach that scales well \textit{and} retrieves high-quality patterns.
While other approaches show good performance on individual datasets, e.g. \tree on \textit{Cardio} and \premise on \textit{Disease} data, they fail to do so in general. We also note that \tree and \spumante tend to return thousands of patterns, which undermines the goal of human interpretation.

\ourmethod fulfills the goal we set for this work and presents itself as a suitable candidate to take on the challenge of high-dimensional pattern mining in applications like genomics. As encouraging its ability in retrieving class-descriptive patterns at scale is, there is of course no free lunch.
For example, on low-dimensional data of up to a hundred features, \ourmethod has a harder time differentiating classes and individual patterns and performs `only' on par with other approaches. %
For such low-dimensional regimes, employing methods with guarantees, that are usually infeasible for large-scale data is still preferential.

Similar to most existing work, \ourmethod considers only conjunctions of features as patterns. In many applications,  relations can be more complex, such as mutually exclusive features. It would make for engaging future work to study extensions of \ourmethod to capture such relations.
In a case study on breast cancer datasets, we show that \ourmethod discovers patterns that capture class-relevant biological processes.
The results are not only encouraging, but also contain many patterns for which the genes are not yet annotated to a pathway or process, or the function of individual genes is still unknown.
These results offer an exciting opportunity to investigate novel links between genes and diseases in follow-up studies with domain experts.

	\section{Conclusion}\label{sec:conclusion}
We studied the problem of discovering differential patterns, i.e., patterns that succinctly describe and 
differentiate between the classes present in the data. Existing methods are often limited to binary classes, do not scale to high-dimensional data, or retrieve uninformative pattern sets.

To tackle this problem, we proposed a novel neural network architecture \ourmethod consisting of a binary autoencoder and a classification head. 
With a flat, binary architecture, the learned intermediate layer captures symbolic patterns.
For the optimization, we proposed a multi-task objective to jointly optimize the reconstruction and classification, thus driving learning of patterns that both reconstruct the data well and differentiate between classes.

On synthetic and real-world data, including biological case studies on breast cancer, we show that \ourmethod strikes a unique balance among existing work, scales to high-dimensional data, is robust to noise, and accurately retrieves differential patterns that are highly interpretable.

	\section*{Acknowledgements}
	The \emph{BRCA} datasets were derived from data made available by the TCGA Research Network.\!\footnote{\url{https://www.cancer.gov/tcga}} 
	
	\bibliography{bib/abbreviations,bib/bib-jilles,bib/bib-paper}
	
	\ifapx
	\appendix
	\newpage
\section{Appendix}
\label{sec:apx}
\subsection{Clamping Function}
The clamping function used in section \ref{sec:forward} is given by
$$
\operatorname{clamp}(x, a, b)= \begin{cases}a & \text { if } x<a, \\ x & \text { if } a \leq x \leq b, \\ b & \text { if } b<x \; .\end{cases}
$$
The $W$-shaped regualrizer is plotted in Fig. \ref{fig:regu} for $\lambda=\kappa=1$.
After each epoch, $\lambda$ and $\kappa$ are increased using an exponential scheduler \cite{dalleiger2022efficiently}.
\begin{figure}[!h]
\hspace{0.75cm}
    \begin{tikzpicture}
        \usetikzlibrary{calc}
        \begin{axis}[ 
            jonas line,
            legend style={at={(0.45, 1.1)}, anchor=south},
            width = 7cm,
            height=4cm,
            ylabel={$r_b(x)$},
            xlabel={$w$},
            ymax=2.0,
            legend columns = 6,
            tick label style={/pgf/number format/fixed}
        ]
             \addplot[color=black] coordinates{
                    ( -1.0 , 2.0 )
                    ( -0.9 , 1.71 )
                    ( -0.8 , 1.44 )
                    ( -0.7 , 1.19 )
                    ( -0.6 , 0.96 )
                    ( -0.5 , 0.75 )
                    ( -0.4 , 0.56 )
                    ( -0.3 , 0.39 )
                    ( -0.2 , 0.24 )
                    ( -0.1 , 0.11 )
                    ( -0.0 , 0.0 )
                    ( 0.1 , 0.11 )
                    ( 0.2 , 0.24 )
                    ( 0.3 , 0.39 )
                    ( 0.4 , 0.56 )
                    ( 0.5 , 0.75 )
                    ( 0.6 , 0.56 )
                    ( 0.7 , 0.39 )
                    ( 0.8 , 0.24 )
                    ( 0.9 , 0.11 )
                    ( 1.0 , 0.0 )
                    ( 1.1 , 0.11 )
                    ( 1.2 , 0.24 )
                    ( 1.3 , 0.39 )
                    ( 1.4 , 0.56 )
                    ( 1.5 , 0.75 )
                    ( 1.6 , 0.96 )
                    ( 1.7 , 1.19 )
                    ( 1.8 , 1.44 )
                    ( 1.9 , 1.71 )
                    ( 2.0 , 2.0 )
             };
        \end{axis}
      \end{tikzpicture}
      \caption{\textit{$W$-shaped regularizer.} To push the weights towards a binary solution during optimization, which allows
               a better quantization, we employ the $W$-shaped regularizer discussed in the main paper (here, $\lambda = \kappa = 1$). }
      \label{fig:regu}
\end{figure}
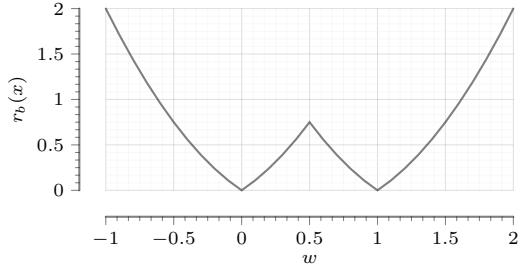
\subsection{Method Details}
\subsubsection{Extracting Differential Patterns}
\label{app:extracting}
The differential patterns are extracted in two steps: First, we extract all patterns
encoded in the pattern layer. Next, we assign the extracted patterns to the corresponding
differential pattern sets $P^k$.

Given the trained continuous weights of the autoencoder $W^E \in [0,1]^{h\times m}$ and weights of
the classifier $W^C \in [0,1]^{K\times h}$, we start by extracting all patterns $P$ encoded in the pattern layer. For this,
we binarize the weight of the autoencoder with a fixed threshold $\tau_{e}$. The pattern $p_i$ encoded in
the $i$-th neuron  is given by $p_i = \{j\mid W^E[i,j]>\tau_{e}, ~j \in \{1...m\} \}$, for $i\in \{1,...,h\}$. The overall
pattern set is given by $P = \{p_i \mid  p_i \neq \emptyset, i\in \{1,...,h\} \}$.
Next, we assign the differential patterns to the classes,
for which we use a threshold $\tau_c$ to binarize the weights of the classifier.
The differential patterns $P^k$ for class $k \in \{1,...,K\}$ are given by
$$P^k=\{p_i \mid W^C[k,i]>\tau_c, ~ p_i \in P \} \;.$$
Informally, a pattern $p_i$ is assigned to the differential pattern set $P_k$, if it is connected
to the output for the classification of class $k$. 
In practice, the thresholds can be chosen by grid search and choosing the pair $(\tau_{t_e},\tau_{t_c})$,
for which the discretized network achieves the lowest reconstruction and classification error.

\subsection{Experimental Details}
\label{app:expdetails}

\subsubsection{Hardware for Experiments}
We implemented \ourmethod in PyTorch, and use the publicly available implementations of other methods.  
Those that leverage a GPU, i.e. \ourmethod and \rrl, were run on machines with NVIDIA DGX A100 and AMD Rome 7742 CPUs. 
The others were run on Intel Xeon(R) Gold 6244 machines with 256GB RAM. 
Individual experiments were stopped after 24 hours or if they exceeded 256GB of RAM.

\subsubsection{Hyperparameter Optimization}
Hyperparameters are optimized as follows.
For \ourmethod, we fine-tune reconstruction loss and classification accuracy on a hold-out set.
For \cart, we set the maximal depth to 20 to facilitate reasonable pattern sizes while not harming performance, and optimize Gini impurity.
For \spumante, we mine the top 1 million patterns using a significance threshold  $\alpha =0.05$, a correction term $\gamma = 0.01$, and set the sampling rate to $25\%$ to keep it from running out of memory.
For \classy, we use a beam width of 200 and a maximum search depth of 20, which provides a good tradeoff between pattern-length and computational burden. 
For \rrl, we optimized the number of hidden layers, the number of neurons in the hidden layer,
the learning rate, and weight decay based on the performance on a hold-out set.

\subsection{Synthetic Data}
\subsubsection{Formal Definition of Soft F1 Score}
\label{app:f1}
To avoid over-penalization of methods that  only recover sub-parts of the individual patterns,
we adopt the soft F1 score from \citet{hedderich2022label}.
Instead of using a strict equality for computing recall and precision, we resort to using
Jaccard distance. Formally, we define the soft F1 score as 
$$
\begin{aligned}
\operatorname{SoftPrec}\left(P_d, P_g\right) & = \frac{1}{|P_d|}\sum_{p_d \in P_d} \underset{p_g \in P_g}{\operatorname{max}} \frac{\left|p_d \cap p_g\right|}{\left|p_d \cup p_g\right|} \; , \\
\operatorname{SoftRec}\left(P_d, P_g\right) & = \frac{1}{|P_g|}  \sum_{p_g \in P_g} \underset{p_d \in P_d}{\operatorname{max}} \frac{\left|p_d \cap p_g\right|}{\left|p_d \cup p_g\right|} \; , \\
\operatorname{SoftF1}\left(P_d, P_g\right) & =\frac{2 * \operatorname{SoftPrec} * \operatorname{ SoftRec }}{\operatorname { SoftPrec }+ \operatorname { SoftRec }} \; ,
\end{aligned}
$$
where we denote the sets of ground truth resp. discovered patterns by $P_g$ and $P_d$.

\subsubsection{Additional Results}
Here, we report additional statistics and results for the experiment on synthetic data.
In Fig. \ref{fig:1b-prec} and Fig. \ref{fig:1b-recall}, we report the precision and recall  for
the scalability in $m$ (Sec. \ref{sec:scam}).
\ourmethod performs equally well with regard to precision and recall.
In contrast, our competitors achieve a significantly higher recall than precision.
This is especially prevalent for \premise and \cart, which explains the overall low F1-score in Fig. \ref{fig:exp1b}.

In Fig. \ref{app:fig-noise}, we report the F1-score for varying numbers of random additive features $a \in \{ 0,10,20,...,90,100\}$. On average,
\ourmethod outperforms all competitors, while \premise has an overall smaller variance. \spumante
suffers from large variance and degrades after 50 random features and runs out of memory $a=100$.

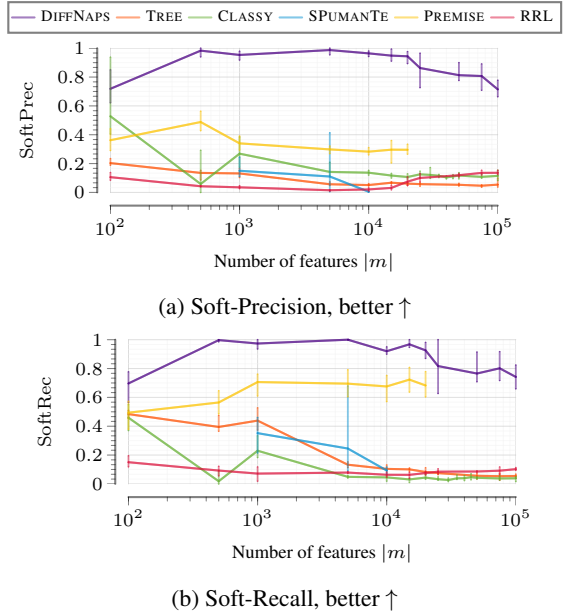
\begin{figure}[ht]

    \begin{subfigure}[b]{\linewidth}
    \centering
        \begin{tikzpicture}
                \usetikzlibrary{calc}
                \begin{semilogxaxis}[ 
                appendix line,
                legend style={at={(0.45, 1.1)}, anchor=south,font={\tiny\arraycolsep=1pt}},
                width = 0.8\linewidth,
            height=3.5cm,
                ylabel={$\operatorname{SoftPrec}$},
                xlabel={Number of features $|m|$},
                ymax=1.0,
                legend columns = 7,
                ]
        
                \foreach \x in {diffnaps, tree,  mdlrl, spumante, premise,rrl}{
                       \barminmaxplot{\x }{SP}{exp1};
                }
    
                \addlegendentry{\ourmethod}
                \addlegendentry{\textsc{Tree}}   
                \addlegendentry{\textsc{Classy}}
                \addlegendentry{\textsc{SPumanTe}} 
                \addlegendentry{\textsc{Premise}}  
                \addlegendentry{\textsc{RRL}}  
                \addlegendentry{\textsc{Subgroup}} 
              
                \end{semilogxaxis}
              \end{tikzpicture}
          \caption{Soft-Precision, better $\uparrow$}
          \label{fig:1b-prec}
        \end{subfigure}
        
        \begin{subfigure}[b]{\linewidth}
        \centering

          \begin{tikzpicture}
            \usetikzlibrary{calc}
            \begin{semilogxaxis}[ 
            appendix line,
            legend style={at={(0.45, 1.1)}, anchor=south,font={\tiny\arraycolsep=1pt}},
            width = 0.8\linewidth,
            height=3.5cm,
            ylabel={$\operatorname{SoftRec}$},
            xlabel={Number of features $|m|$},
            ymax=1.0,
            legend columns = 7,
            ]
    
            \foreach \x in {diffnaps, tree,  mdlrl, spumante, premise,rrl}{
                   \barminmaxplot{\x }{SR}{exp1};
            }

            \end{semilogxaxis}
          \end{tikzpicture}
          \caption{Soft-Recall, better $\uparrow$}
          \label{fig:1b-recall}      
        \end{subfigure}
    \caption{\textit{Additional results for Scalability.} We provide the soft precision score \textbf{(a)} and soft recall
            score \textbf{(b)} on the synthetic data with an increasing number of features. }

\end{figure}

\begin{figure}[h]
    \centering
    \begin{tikzpicture}
        \usetikzlibrary{calc}
        \begin{axis}[ 
            appendix line,
            legend style={at={(0.45, 1.1)}, anchor=south,font={\tiny\arraycolsep=1pt}},
            width = 0.8\linewidth,
            height=3.5cm,
            ylabel={Soft-F1},
            xlabel={$a$},
            ymax=1.0,
            legend columns = 7,
            xmin=0.0,
        ]
        \foreach \x in {diffnaps, tree,  mdlrl, spumante, premise,rrl}{
               \barminmaxplot{\x }{F1}{exp3};
        }
            \addlegendentry{\ourmethod}
            \addlegendentry{\textsc{Tree}}   
            \addlegendentry{\textsc{Classy}}
            \addlegendentry{\textsc{SPumanTe}} 
            \addlegendentry{\textsc{Premise}}  
            \addlegendentry{\textsc{RRL}}  
        \end{axis}
      \end{tikzpicture}
      \caption{\emph{Results for additive noise} 
      We report the soft F1 score on synthetic data with different amounts of 
      random additive features  $a \in \{ 0,10,20,...,90,100\}$.
      }
      \label{app:fig-noise}
      \label{fig:exp3}
\end{figure}
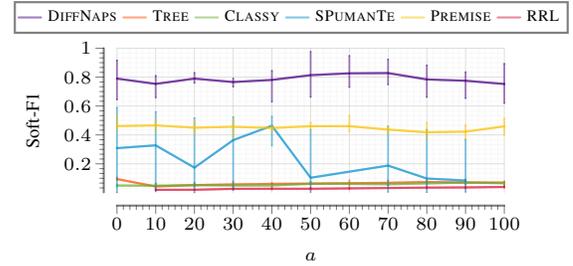

\subsection{Real data}

\begin{table*}
    \centering \fontsize{9}{11}\selectfont
    \begin{tabular}{ l r r r r r r r r } 
        \toprule
        \multicolumn{4}{c}{} & \multicolumn{5}{c}{Runtime} \\
        \cmidrule(l){5-9}
        Dataset & Rows & Columns & $k$ &\ourmethod (ours)  & \cart & \classy & \textsc{Premise} & \spumante   \\
        \midrule
            Cardio     & 68k & 45     & 2  & 1m33s & 1s & 15s & 10s & 43s\\
            Disease  & 5k  & 131    & 41 & 1m40s & 1s & 14s & 8s & 48s\\
            BRCA-N            & 222   & 20k  & 2m & 18s & 1s & 33m40s & \na & 3h45m\\
            BRCA-S            & 187   & 20k  & 4m & 58s & 1s & 26m08s & \na & 2h31m\\
            Genomes       & 2.5k  & 225k & 6 & 8m59s & 28s & 8h20m & \na & \na\\
        \bottomrule
        \end{tabular}
        \caption{\textit{Runtime real world data.} We specify the time taken to compute the results. 
        We abort experiments that ran out of memory or took longer than 24h (\na).}\label{app:tab-runtime}
\end{table*}
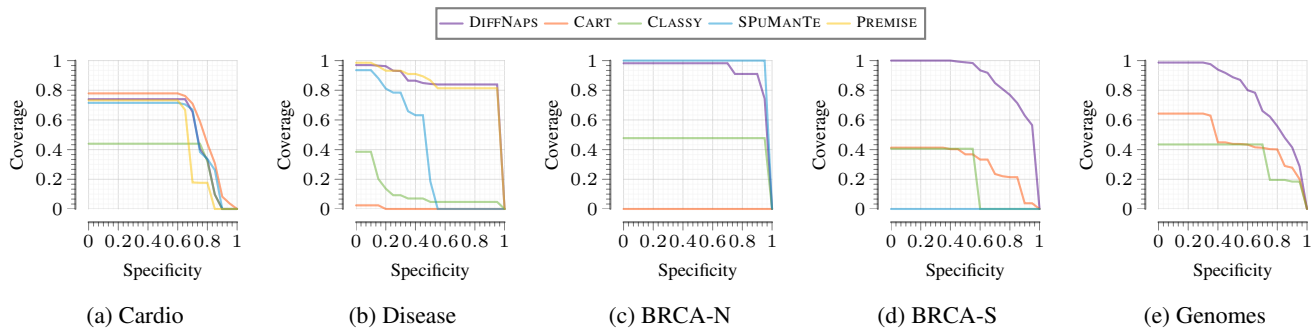
\begin{figure*}[ht]
    \centering
    \captionsetup[subfigure]{justification=centering}
    \begin{subfigure}[b]{0.2\textwidth}
        \begin{tikzpicture}
        \usetikzlibrary{calc}
            \begin{axis}[ 
                roc line,
                legend style={xshift=9.2cm,yshift=0.75cm},
                xlabel={Specificity},
                ylabel={Coverage},
                ymax=1.0,
                xmax=1.0,
                legend columns = 6,
                tick label style={/pgf/number format/fixed}
            ]
            \foreach \x in {diffnaps, tree,  mdlrl, spumante, premise}
                \addplot table[x=x,y=cardio-\x,col sep=comma]{results/roc_curve.csv};
            \addlegendentry{\ourmethod}
            \addlegendentry{\tree}
            \addlegendentry{\classy}
            \addlegendentry{\spumante}
            \addlegendentry{\premise}
            \end{axis}
       \end{tikzpicture}
       \subcaption{Cardio}
    \end{subfigure}%
    \begin{subfigure}[b]{0.2\textwidth}
        \begin{tikzpicture}
        \usetikzlibrary{calc}
            \begin{axis}[ 
                roc line,
                legend style={at={(0.45, 1.1)}, anchor=south,font={\tiny\arraycolsep=1pt}},
                xlabel={Specificity},
                ylabel={Coverage},
                ymax=1.0,
                xmax=1.0,
                legend columns = 5,
                tick label style={/pgf/number format/fixed}
            ]
            \foreach \x in {diffnaps, tree,  mdlrl, spumante, premise}
                \addplot table[x=x,y=disease-\x,col sep=comma]{results/roc_curve.csv};
            \end{axis}
       \end{tikzpicture}
     \subcaption{Disease}
    \end{subfigure}%
    \begin{subfigure}[b]{0.2\textwidth}
        \begin{tikzpicture}
        \usetikzlibrary{calc}
            \begin{axis}[ 
                roc line,
                legend style={at={(0.45, 1.1)}, anchor=south,font={\tiny\arraycolsep=1pt}},
                xlabel={Specificity},
                ylabel={Coverage},
                ymax=1.0,
                xmax=1.0,
                legend columns = 5,
                tick label style={/pgf/number format/fixed}
            ]
            \foreach \x in {diffnaps, tree,  mdlrl, spumante}
                \addplot table[x=x,y=brca-\x,col sep=comma]{results/roc_curve.csv};
            \end{axis}
       \end{tikzpicture}
       \subcaption{BRCA-N}
    \end{subfigure}%
    \begin{subfigure}[b]{0.2\textwidth}
        \begin{tikzpicture}
        \usetikzlibrary{calc}
            \begin{axis}[ 
                roc line,
                legend style={at={(0.45, 1.1)}, anchor=south,font={\tiny\arraycolsep=1pt}},
                xlabel={Specificity},
                ylabel={Coverage},
                ymax=1.0,
                xmax=1.0,
                legend columns = 5,
                tick label style={/pgf/number format/fixed}
            ]
            \foreach \x in {diffnaps, tree,  mdlrl, spumante}
                \addplot table[x=x,y=brca_mult-\x,col sep=comma]{results/roc_curve.csv};
            \end{axis}
       \end{tikzpicture}
       \subcaption{BRCA-S}
    \end{subfigure}%
    \begin{subfigure}[b]{0.2\textwidth}
        \begin{tikzpicture}
        \usetikzlibrary{calc}
            \begin{axis}[ 
                roc line,
                legend style={at={(0.45, 1.1)}, anchor=south,font={\tiny\arraycolsep=1pt}},
                xlabel={Specificity},
                ylabel={Coverage},
                ymax=1.0,
                xmax=1.0,
                legend columns = 5,
                tick label style={/pgf/number format/fixed}
            ]
            \foreach \x in {diffnaps, tree,  mdlrl}
                \addplot table[x=x,y=genomes-\x,col sep=comma]{results/roc_curve.csv};
            \end{axis}
       \end{tikzpicture}
       \subcaption{Genomes}
    \end{subfigure}%
    \caption{ \textit{Real world data.} Here, we plot the specificity-coverage curves used to compute the AUCs reported in Tab. \ref{tab:real}.
    The curves are computed as explained in Sec. \ref{app:auc}. }
    \label{fig:odd-coverage}
\end{figure*}
\subsubsection{Computing AUCs}
\label{app:auc}
As we do not have the ground truth for real world data, we resort to evaluating the area under the curve of what percentage of the data the patterns cover when we order them by the probability of seeing a class given a pattern.
This can be roughly translated into sensitivity (how much of the dataset do we cover) versus specificity
(how specific is the pattern for that class). To filter spurious patterns, 
we only consider patterns with a predictive probability $\frac{1}{K}+0.1$ (i.e., at least slightly more likely than chance).
More formally on the x-axis we plot
$\Pro(k \mid p_k)=\frac{\mathit{supp}_k(p_k) }{\mathit{supp}(p_k)}\;.$
If a pattern  $p_k$is very specific to a class $k$, then $\Pro(k \mid p_k) = 1$ and
if it is unspecific $\Pro(k \mid p_k) = 0$

On the y-axis, we plot how much of the dataset is covered (explained)
given all the patterns that pass the threshold $p$ and have a predictive probability $\frac{1}{K}+0.1$ .
The coverage $\operatorname{cov}(p,X) $ of a pattern $p$ for a dataset $X$ is defined as
$$ \operatorname{cov}(p,X)  = \{ X_i \mid p \subset X_i \} \; .$$
With a slight abuse of notation, the coverage of a pattern set $\mathcal{P}$ is then given by
$$ \operatorname{cov}(\mathcal{P},X)  = \bigcup_{p\in \mathcal{P}} \operatorname{cov}(p,X) \; .$$
With  $P^k_q$, we denote the set of patterns such that
such that $\Pro(k \mid p_k) > p $ and is not spurious. Then for a threshold
$p$, the corresponding value on the y-axis $y_p$ is computed as 
$$y_p = \frac{1}{K} \sum_{k=1}^K \frac{ | \operatorname{cov}( P^k_q, X^k)| }{|X^k|} \; .$$
That is, per class $k$, we compute how much of the corresponding partition is covered and take the mean of 
those individual coverages.

\subsubsection{Sensitivity-Specificity Curves}
Figure \ref{fig:odd-coverage} shows the sensitivity-specificity curves for the AUCs in (Table \ref{tab:real}).

\subsubsection{Processing of BRCA Data}
\label{app:brca-gen}

We obtained re-aligned bulk RNA-seq data of TCGA BRCA samples through recount3~\cite{recount3}.
We first filtering samples into primary tumor and adjacent normal tissue samples and keeping only protein-coding genes with non-zero expression in at least one sample. We then remove duplicate samples of individuals, keeping the one with highest sequencing depth.
Gene expression counts were log-TPM transformed.

To binarize the expression data, for each gene, we set samples that have expression larger than the upper quartile to 1, all others to 0. If the upper quartile is 0, we set all non-zero samples to 1.

For the normal vs tumor data (\textit{BRCA-N}), as many of the competitors are sensitive to class-imbalance, we kept only the matching samples, i.e., where individuals where both adjacent normal as well as primary tumor tissue was available.

For the BRCA subtype data (\textit{BRCA-S}), we followed the same simple binarization scheme, and kept at most $50$ samples per subtype to keep the data roughly balanced, sampling at random without replacement.
The four subtypes---luminal A, luminal B, HER2+, and triple negative---where defined based on annotated receptor status of the Estrogen recpetor (ER), the Progesterone receptor (PR), and the human epidermal growth factor receptor 2 (HER2) available from the recount data.
In particular, we define luminal A as (ER+, PR+, HER2-), luminal B as (ER+, PR-, HER2-), HER2+ as obvious, and triple negative as (ER-, PR-, HER2-). We removed all samples that do not belong to any of these subtypes or where receptor status was not available.

\subsubsection{Analysis of BRCA Patterns}
\label{app:brca-ana}

To analyze pattern sets discovered by \ourmethod qualitatively in terms of whether they represent reasonable biological functions specific to a label, we compute gene set over-representation
statistics for gene relationships annotated in the Kyoto Encyclopedia of Genes and Genomes (KEGG).
KEGG serves as a gold standard for known biological pathways and relationships, including hand-drawn and manually curated cellular pathways.
A gene set over-representation analysis tests whether an overlap of a given gene set (e.g., a pattern or union of patterns) with an annotated pathway is more likely than chance, where the null overlap statistic is computed using a background gene set (here: the set of genes in the dataset).
We use the \textsc{enrichR} software package for the gene set over-representation analysis and report results as significant with a p-value cutoff of $.05$~\cite{enrichr}.
For an overall assessment we consider pathways that are found for the union of genes across all patterns for a class. We also enriched pathways for each pattern individually, many of the patterns, however, were too small to be considered for enrichment or contained genes for which no annotation  is available in KEGG.
To obtain an estimate of the average log-odds ratios of likelihood of a pattern set $P$ describing a class, we compute
$ \frac{1}{|P|}\sum_{p\in P}ln\left(\frac{P(p\mid l)}{P(p\mid \lnot k)}\right)$, where $ln$ is the
natural logarithm and we do not compute those terms where $P(p\mid \lnot k)=0$, which leaves us with a lower bound of the log-odds.

	\fi
	
\end{document}